\documentclass[12pt]{article}

\usepackage{aaai}
\usepackage{epsfig}
\usepackage{wrapfig}
\usepackage{latexsym}
\usepackage{fancybox}
\usepackage{wrapfig}

\begin{document}
\title{\ccgolog: Towards More Realistic Logic-Based Robot Controllers}

\author{{\bf Henrik Grosskreutz} and {\bf Gerhard Lakemeyer}\\[1ex]
Department of Computer Science V\\
Aachen University of Technology\\
52056 Aachen, Germany\\
\{grosskreutz,gerhard\}@cs.rwth-aachen.de}

\newcommand{\ccgolog}{{\sf cc-Golog}}
\newcommand{\congolog}{{\sf ConGolog}}

\maketitle

\begin{abstract}
  
  High-level robot controllers in realistic domains typically deal with
  processes which operate concurrently, change the world continuously, and
  where the execution of actions is event-driven as in ``charge the batteries
  as soon as the voltage level is low''. While non-logic-based robot control
  languages are well suited to express such scenarios, they fare poorly when
  it comes to projecting, in a conspicuous way, how the world evolves when
  actions are executed. On the other hand, a logic-based control language like
  \congolog, based on the situation calculus, is well-suited for the latter.
  However, it has problems expressing event-driven behavior. In this paper, we
  show how these problems can be overcome by first extending the situation
  calculus to support continuous change and event-driven behavior and then
  presenting \ccgolog, a variant of \congolog\ which is based on the extended
  situation calculus.  One benefit of \ccgolog\ is that it narrows the gap in
  expressiveness compared to non-logic-based control languages while
  preserving a semantically well-founded projection mechanism.

\end{abstract}

\bibliographystyle{aaai}

\newcommand{\rc}[1]{{\scriptsize {\sf #1}}}
\newcommand{\rpl}{{\sc rpl}}
\newcommand{\golog}{{\sf GOLOG}}
\newcommand{\srp}{{\sc srp}}
\newcommand{\xfrm}{{\sc xfrm}}
\newcommand{\xfrmml}{{\sc xfrm-ml}}
\newcommand{\rplcode}[1]{{\footnotesize {\sf #1}}}
\newcommand{\rcit}[1]{\rc{{\it #1}}}
\newcommand{\rhino}{{\sc Rhino}}
\newcommand{\carl}{{\sf CARL}}
\newcommand{\faxbot}{{\sc Faxbot}}
\newcommand{\ep}{{$e{\rightarrow}p$}}
\newcommand{\pe}{{$p{\rightarrow}e$}}
\newcommand{\pp}{{$p{\leftarrow}p$}}
\newcommand{\true}{{\it true}}
\newcommand{\false}{{\it false}}
\newcommand{\step}{{$step$}}
\newcommand{\steps}{{$steps$}}
\newcommand{\change}{{$change$}}
\newcommand{\changes}{{$changes$}}
\newcommand{\R}{{\mathbb{R}}}
\newcommand{\ia}{{\bf I.A.:}}
\newcommand{\is}{{\bf I.S.:}}

\gdef\M#1{\ifmmode #1\else$#1$\fi}
\newcommand{\code}[1]{\textbf{#1}}%
\newcommand{\name}[1]{\textit{#1}}%
\newcommand{\nameX}[2]{\name{#1}\code{(}\name{#2}\code{)}}%
\newcommand{\nameXX}[3]{\nameX{#1}{#2\code{,} #3}}%

\newcommand{\ldef}{=^{def}}

\gdef\fvar#1{\ifmmode \sf{#1}\else$\sf{#1}$\fi}

\newcommand{\bi}{\sl}

\gdef\M#1{\ifmmode #1\else$#1$\fi}

\newcommand{\entails}{\M\models}

\newcommand{\limp}{\M{\supset}}
\renewcommand{\land}{\M{\wedge}}
\renewcommand{\lor}{\M{\vee}}
\renewcommand{\lnot}{\M{\neg}}

\newcommand{\Szero}{\M{S_0}}
\newcommand{\Poss}{\mbox{\it Poss}}

\newcommand{\SF}{\mbox{\it SF}}

\newcommand{\Knows}{{\tt Knows}}

\newcommand{\Kwhether}{{\tt Kwhether}}
\newcommand{\Kref}{{\tt Kref}}

\newcommand{\now}{{\it now}}
\newcommand{\AX}{{\rm AX}}

\newcommand{\dofn}{\mbox{\it do}}

\newcommand{\aone}{\M{a_1}}
\newcommand{\atwo}{\M{a_2}}
\newcommand{\athree}{\M{a_3}}
\newcommand{\afour}{\M{a_4}}
\newcommand{\afive}{\M{a_5}}
\newcommand{\cone}{\M{c_1}}
\newcommand{\ctwo}{\M{c_2}}
\newcommand{\cthree}{\M{c_3}}
\newcommand{\cpone}{\M{c_1'}}
\newcommand{\cptwo}{\M{c_2'}}
\newcommand{\cpthree}{\M{c_3'}}

\newcommand{\sitfree}{\mbox{situation-free}}

\newcommand{\Do}{\mbox{\it Do}}
\newcommand{\DoOld}{\mbox{\it Do$_{\mbox{\begin{tiny}Old\end{tiny}}}$}}
\newcommand{\poss}{\mbox{\it Poss}}
\newcommand{\branchon}{\M{branch\_on}}
\newcommand{\cdo}{\M{cdo}}
\newcommand{\cat}{\M{\cdot}}
\newcommand{\ext}{\M{ext}}
\newcommand{\exttwo}{\M{ext'}}
\newcommand{\Dof}{\M{Do4}}
\newcommand{\catchplane}{\mbox{\it catch\_plane}}
\newcommand{\boardplane}{\mbox{\it board\_plane}}
\newcommand{\sensegate}{\mbox{\it sense\_gate}}
\newcommand{\buycoffee}{\mbox{\it buy\_coffee}}
\newcommand{\buypaper}{\mbox{\it buy\_paper}}
\newcommand{\gotogate}{\mbox{\it goto\_gate}}
\newcommand{\goto}{\mbox{\it goto}}
\newcommand{\atairport}{\mbox{\it at\_airport}}
\newcommand{\gate}{\mbox{\it gate}}
\newcommand{\gateA}{\mbox{\it gate\_A}}
\newcommand{\gateB}{\mbox{\it gate\_B}}
\newcommand{\amat}{\mbox{\it am\_at}}
\newcommand{\amatgate}{\mbox{\it am\_at\_gate}}
\newcommand{\itisgateA}{\mbox{\it it\_is\_gate\_A}}
\newcommand{\sense}{\mbox{\it sense}}
\newcommand{\sensed}{\mbox{\it sensed}}
\newcommand{\assm}{\mbox{\it assm}}
\newcommand{\Holds}{\mbox{\it Holds}}
\newcommand{\PhiFplus}{\M{\Phi_F^+}}
\newcommand{\PhiFminus}{\M{\Phi_F^-}}
\newcommand{\rhoone}{\M{\rho_1}}
\newcommand{\rhotwo}{\M{\rho_2}}
\newcommand{\path}{\M{path}}
\newcommand{\eps}{\M{\epsilon}}

\def \marg#1{\marginpar{\tiny #1}}
\newcommand{\nl}[1]{\mbox{\it #1}}      
\newcommand{\waitfor}{\nl{waitFor}}
\newcommand{\waitFor}{\nl{waitFor}}
\newcommand{\startGo}{\nl{startGo}}
\newcommand{\stopGo}{\nl{endGo}}
\newcommand{\robotLoc}{\nl{robotLoc}}
\newcommand{\Time}{\nl{time}}
\newcommand{\linear}{\nl{linear}}
\newcommand{\constant}{\nl{constant}}
\newcommand{\withPol}{\nl{withPol}}
\newcommand{\tryAll}{\M{tryAll}}
\newcommand{\para}{\M{par}}
\newcommand{\nextMorning}{\nl{nextMorning}}
\newcommand{\nextEvening}{\nl{nextEvening}}
\newcommand{\nextMorn}{\nl{nextMorn}}
\newcommand{\nextEve}{\nl{nextEve}}
\newcommand{\say}{\nl{say}}
\newcommand{\setTime}{\nl{setTime}}
\newcommand{\lowVltg}{\nl{lowVltg}}
\newcommand{\setLowVltg}{\nl{setLowVltg}}
\newcommand{\chargeBatteries}{\nl{chargeBatteries}}
\newcommand{\delivered}{\nl{delivered}}
\newcommand{\mail}{\nl{mail}}
\newcommand{\loc}{\nl{loc}}
\newcommand{\dest}{\nl{dest}}
\newcommand{\deliver}{\nl{deliver}}
\newcommand{\speed}{\nl{speed}}
\newcommand{\startGoAtSpeed}{\nl{startGoAtSpeed}}
\newcommand{\atDesk}{\nl{atDesk}}
\newcommand{\leaveRoom}{\nl{leaveRoom}}
\newcommand{\coffee}{\nl{coffee}}
\newcommand{\informSupervisor}{\nl{informSupervisor}}
\newcommand{\give}{\nl{give}}
\newcommand{\enterRoom}{\nl{enterRoom}}
\newcommand{\hallway}{\nl{hallway}}
\newcommand{\office}{\nl{office}}
\newcommand{\start}{\nl{start}}
\renewcommand{\Poss}{\nl{Poss}}
\newcommand{\vel}{\nl{v}}
\newcommand{\val}{\nl{val}}
\newcommand{\ltp}{\nl{ltp}}
\newcommand{\runBackup}{\nl{runBackup}}
\newcommand{\clock}{\nl{clock}}

\newcommand{\rap}{{\sc rap}}
\newcommand{\colbert}{{\sc colbert}}

\newtheorem{THEOREM}{Theorem}
\newtheorem{theorem}{Theorem}[section]
\newtheorem{LEMMA}[THEOREM]{Lemma}
\newenvironment{lemma}{\begin{LEMMA} \hspace{-.85em} {\bf :} }%
                      {\end{LEMMA}}
\newtheorem{COROLLARY}[THEOREM]{Corollary}
\newenvironment{corollary}{\begin{COROLLARY} \hspace{-.85em} {\bf :} }%
                          {\end{COROLLARY}}
\newtheorem{PROPOSITION}[THEOREM]{Proposition}
\newenvironment{proposition}{\begin{PROPOSITION} \hspace{-.85em} {\bf :} }%
                            {\end{PROPOSITION}}
\newtheorem{DEFINITION}[THEOREM]{Definition}
\newenvironment{definition}{\begin{DEFINITION} \hspace{-.85em} {\bf :} \rm}%
                            {\end{DEFINITION}}
\newtheorem{CLAIM}[THEOREM]{Claim}
\newenvironment{claim}{\begin{CLAIM} \hspace{-.85em} {\bf :} \rm}%
                            {\end{CLAIM}}
\newtheorem{EXAMPLE}[THEOREM]{Example}
\newenvironment{example}{\begin{EXAMPLE} \hspace{-.85em} {\bf :} \rm}%
                            {\end{EXAMPLE}}
\newtheorem{REMARK}[THEOREM]{Remark}
\newenvironment{remark}{\begin{REMARK} \hspace{-.85em} {\bf :} \rm}%
                            {\end{REMARK}}

\newcommand{\tfunction}{{\rm t-function}}
\newcommand{\tform}{{\rm t-form}}

\section{Introduction}

High-level robot controllers typically specify processes which operate
concurrently and change the world in a continuous fashion over time.
Several special programming languages such as
\rpl~\cite{McD92Rob},\footnote{\rpl\ has recently been used successfully to
control the behavior of a mobile robot deployed in a realistic environment
for an extended period of time~\cite{Thr99Min}.}  \rap~\cite{Fir87RAP}, or
\colbert~\cite{Kon97Col} have been developed for this purpose.  As an
example, consider the following \rpl-program:

\begin{figure}[htbp]
\vspace*{-2ex}
  \begin{center}
      \parbox{5cm}{%
        \begin{scriptsize}
          \begin{sf}
            \begin{tabbing}
              $~~$ \underline{WITH-POLICY} \=
              \underline{WHENEVER} Batt-Level $\le$ 46 $~~$\\
              \> \hspace{2ex} CHARGE-BATTERIES \\
              $~~~~~$ \underline{WITH-POLICY}  \underline{WHENEVER}
              NEAR-DOOR(RmA-118) \\
              \> \hspace{3ex} \= SAY(``hello'') \\
              $~~~~~~~$ DELIVER-MAIL \\
            \end{tabbing}
          \end{sf}
        \end{scriptsize}}
  \end{center}

  \vspace{-6ex}
  
  \caption{Office delivery plan}
  \label{fig:office-delivery-plan}
  \vspace{-1ex}
\end{figure}

Roughly, the robot's main task is to deliver mail, which we merely indicate by a
call to the procedure {\scriptsize DELIVER-MAIL}. While executing this
procedure, the robot concurrently also does the following, with an increasing
level of priority: whenever it passes the door to Room A-118 it says ``hello''
and, should the battery level drop dangerously low, it recharges its batteries
interrupting whatever else it is doing at this moment.

Even this simple program exhibits important features of high-level robot
controllers: (1) The timing of actions is largely {\em event-driven}, that
is, rather than explicitly stating when an action occurs, the execution
time depends on certain conditions becoming true such as reaching a certain
door. Most robot control languages realize this feature using the special
construct \waitFor($\phi$), which suspends activity until $\phi$ becomes
true.\footnote{In the example, \waitFor\ is hidden within the
whenever-construct.}  (2) Actions are executed {\em as soon as
possible}. For example, the batteries are charged immediately after a low
voltage level is determined.  (3) Conditions such as the voltage level are
best thought of as changing {\em continuously} over time.  (4) Parts of
programs which execute concurrently and with high priority must be {\em
non-blocking}. For example, while waiting for a low battery level, mail
delivery should continue. On the other hand, the actual charging of the
battery should block all other activity.

Given the inherent complexity of concurrent robot programs,
answers to questions like whether a program is executable
and whether it will satisfy the intended goals are not
easy to come by, yet important to both
the designer during program development and the robot who
may want to choose among different courses of action.
A principled approach to answering such
questions is to {\em project} how the world evolves when
actions are performed, a method which also lies at the
heart of planning.

In the case of \rpl, a projection mechanism called \xfrm\ 
exists~\cite{McD92Rob,McD94Alg}, but it has problems.\footnote{As far as we
  know, other non-logic-based robot control languages like \rap\ or \colbert\ 
  do not even consider projection.}  Perhaps the most serious deficiency of
\xfrm\ is that projections rely on using \rpl's execution mechanism, which
lacks a formal semantics and which makes predictions implementation dependent.
Preferably one would like a language which is as powerful as \rpl\ yet allows
for projections based on a perspicuous, declarative semantics.

The recently proposed language \congolog~\cite{Gia97Rea} fulfills
some of these desiderata as it offers many of the features of \rpl\  such
as concurrency, priorities etc. and, at the same time, supports
rigorous projections of plans\footnote{In this paper we will
use the terms program and plan interchangeably, following
McDermott~\cite{McD92Rob} who takes plans to be programs whose execution
can be reasoned about by the agent who executes the program.}  because it
is entirely based on the situation calculus\cite{McC63Sit,Lev98Fou}. 

It turns out, however, that despite many similarities, \congolog\ in its
current form is not suitable to represent robot controllers such as the
example above.  The main problem is that the existing temporal extensions
of the situation calculus such as~\cite{Pin97Int,Rei96Nat} require that the
execution time of an action is supplied explicitly, which seems incompatible
with event-driven specifications. To solve this problem we proceed in
two steps.  First we present a new extension of the situation calculus
which, besides dealing with continuous change, allows us to model actions
which are event-driven by including \waitFor\ as a special action in the
logic. We then turn to a new variant of \congolog\ called \ccgolog, which
is based on the extended situation calculus.  We study issues arising from
the interaction of \waitFor-actions and concurrency and show how
the example-program can be specified quite naturally in \ccgolog\
with the additional benefit of supporting projections firmly grounded in
logic.

The rest of the paper is organized as follows. In the next section, we
briefly review the basic situation calculus. Then we show how to extend it
to include continuous change and time.  After a very brief summary of
\congolog, we present \ccgolog, which takes into account the extended
situation calculus. This is followed by a note on experimental
results and conclusions.

\section{The Situation Calculus}
One increasingly popular language for representing and reasoning about the
preconditions and effects of actions is the situation calculus
\cite{McC63Sit}. We will not go over the language in detail except to note the
following features: all terms in the language are one of three sorts, ordinary
objects, actions or situations; there is a special constant $S_0$ used to
denote the {\em initial situation}, namely that situation in which no actions
have yet occurred; there is a distinguished binary function symbol $\dofn$
where $\dofn(a,s)$ denotes the successor situation to $s$ resulting from
performing the action $a$; relations whose truth values vary from situation to
situation are called relational {\em fluents\/}, and are denoted by predicate
symbols taking a situation term as their last argument; similarly, functions
varying across situations are called functional fluents and are denoted
analogously; finally, there is a special predicate $\Poss(a,s)$ used to state
that action $a$ is executable in situation $s.$

Within this language, we can formulate theories which describe how the world
changes as the result of the available actions. One possibility is a {\em basic
action theory} of the following form \cite{Lev98Fou}:

\begin{itemize}
\item Axioms describing the initial situation, $S_0$.

\item Action precondition axioms, one for each primitive action $a$, 
characterizing $\Poss(a,s)$.

\item 
\begin{sloppypar}
  Successor state axioms, one for each fluent $F$, stating under what
  conditions $F(\vec{x},\dofn(a,s))$ holds as a function of what holds in
  situation $s.$ These take the place of the so-called effect axioms, but also
  provide a solution to the frame problem \cite{Lev98Fou}. 
\end{sloppypar}

\item Domain closure and unique name axioms for the actions.
  
\item Foundational, domain independent axioms~\cite{Lev98Fou}.
  \begin{enumerate}
  \item $\forall P. P(S_0) \land [\forall s\forall a. (P(s) \limp P(\dofn(a,s)))]
        \limp \forall s P(s)$;
  \item $\dofn(a,s) = \dofn(a',s') \limp a = a' \land s = s'$;\footnote{We 
      use the convention that all free
      variables are implicitly universally quantified.}
  \item $\lnot(s \sqsubset S_0)$;
  \item $s \sqsubset \dofn(a,s') \equiv s \sqsubseteq s'$, where $s \sqsubseteq
    s'$ stands for $(s \sqsubset s')\lor (s = s')$.
  \item $s \prec s' \equiv s \sqsubset s' \land \forall a, s^* . 
s \sqsubset do(a,s^*) \sqsubseteq s' \limp Poss(a,s^*)$
  \end{enumerate}
\end{itemize}

The first is a second-order induction axiom ensuring that the only
situations are those obtained from applying $\dofn$ to $S_0$.  The second
is a unique names axiom for situations. $\sqsubseteq$ defines an
ordering relation over situations. Intuitively, $s \sqsubseteq s'$ holds if
$s'$ can be obtained from $s$ by performing a finite number of actions.
Finally, $s \prec s'$ holds when there is a {\em legal} sequence of actions
leading from $s$ to $s'$, where legal means that each action is possible.

\section{Continuous Change and Time}
Actions in the situation calculus cause discrete changes and, in its basic form,
there is no notion of time.  In robotics applications, however, we are faced
with processes such as navigation which cause properties like the robot's
location and orientation to change continuously over time. In order to model
such processes in the situation calculus in a natural way, we add continous
change and time directly to its ontology.

As demonstrated by Pinto and Reiter~\cite{Pin97Int,Rei96Nat}, adding
time is a simple matter. We add a new sort {\em real} ranging over the real
numbers and, for mnemonic reasons, another sort {\em time} ranging over the
reals as well.\footnote{For simplicity, the reals are not axiomatized and
we assume their standard interpretations together with the usual operations
and ordering relations.}  In order to connect situations and time, we add a
special unary functional fluent \start\ to the language with the
understanding that $\start(s)$ denotes the time when situation $s$
begins. We will see later how \start\ obtains its values and, in
particular, how the passage of time is modeled.

A fundamental assumption of the situation calculus is that fluents have a fixed
value at every given situation. In order to see that this assumption still
allows us to model continuous change, let us consider the example of a mobile
robot moving along a straight line in a 1-dimensional world, that is, the
robot's location at any given time is simply a real number. There are two types
of actions the robot can perform, $\startGo(\vel)$, which initiates moving the
robot with speed $\vel$, and \stopGo\ which stops the movement of the robot. Let
us denote the robot's location by the fluent \robotLoc. What should the value of
\robotLoc\ be after executing $\startGo(\vel)$ in situation $s$? Certainly it
cannot be a fixed real value, since the position should change over time as long
as the robot moves. In fact, the location of the robot at any time after
$\startGo(\vel)$ (and before the robot changes its velocity) can be
characterized (in a somewhat idealized fashion) by the function
$x+\vel\times(t-t_0)$, where $x$ is the starting position and $t_0$ the starting
time. The solution is then to take this {\em function of time} to be the value
of \robotLoc.  We call functional fluents whose values are continuous functions
{\em continuous fluents}.

The idea of continuous fluents, which are often called {\em parameters}, is
not new. Sandewall~\cite{San89Com} proposed it when integrating the
differential equations into logic, Galton \cite{Gal90Cri} investigated similar
issues within a temporal logic, and Shanahan considers continuous change in
the event calculus \cite{Sha90Con}. Finally, Miller and
Pinto~\cite{Mil96Cas,Pin97Int} formulate continuous change in the situation
calculus. Here we essentially follow Pinto, in a somewhat simplified form.

We begin by introducing a new sort {\em \tfunction}, whose elements are
meant to be functions of time. We assume that there are only finitely many
function symbols of type \tfunction\ and we require {\em domain closure and
unique names axioms} for them, just as in the case of primitive actions.
For our robot example, it suffices to consider two kinds of \tfunction{s}:
constant functions, denoted by $\constant(x)$ and the special linear
functions introduced above, which we denote as $\linear(x,\vel,t_0)$.

Next we need to say what values these functions have at any particular time
$t$. We do this with the help of a new binary function \val. In the example,
we would add the following axioms:

\begin{enumerate}
      \setlength{\itemsep}{-3pt}
      \setlength{\parsep}{0pt}
      \setlength{\topsep}{-5pt}    
\item [] $\val(\constant(x),t) = x$;
\item [] $\val(\linear(x,\vel,t_0), t) = x + \vel\times(t-t_0)$.
\end{enumerate}

Let us now turn to the issue of modeling the passage of time during a course
of actions.  As indicated in the introduction, motivated by the treatment of
time in robot control languages like \rpl, \rap, or \colbert, we introduce a
new type of primitive action $\waitFor(\phi)$.  The intuituion is as follows.
Normally, every action happens immediately, that is, the starting time of the
situation after doing $a$ in $s$ is the same as the starting time of $s$. The
only exception is $\waitFor(\phi)$: whenever this action occurs, the starting
time of the resulting situation is advanced to the earliest time in the future
when $\phi$ becomes true.  Note that this has the effect that actions always
happen as soon as possible. One may object that requiring that two actions
other than \waitFor\ must happen at the same time is unrealistic.  However, in
robotics applications, actions often involve little more than sending messages
in order to initiate or terminate processes so that the actual duration of
such actions is negligible. Moreover, if two actions cannot happen at the same
time, they can always be separated explicitly using \waitFor.

For the purposes of this paper, we restrict the argument of \waitFor\ to what we
call a \tform, which is a Boolean combination of closed atomic formulas of the
form $(F\ op\ r)$, where $F$ is a continuous fluent with the situation argument
suppressed, $op\in\{<,=\}$,\footnote{We freely use $\le$, $\ge$, or $>$ as
well.}  and $r$ is a term of type real (not mentioning \val).  An example
is $\phi = (\robotLoc \ge 1000)$. To evaluate a \tform\ at a
situation $s$ and time $t$, we write $\phi[s,t]$ which results in a formula
which is like $\phi$ except that every continuous fluent $F$ is replaced by
$\val(F(s),t)$. For instance, $(\robotLoc \ge
1000)[s,t]$ becomes $(\val(\robotLoc(s),t) \ge 1000)$.  For reasons of space we
completely gloss over the details of reifying \tform{s} within the
language\footnote{See, for example,~\cite{Gia99Fou} for details how this can be
done.}  except to note that we introduce \tform{s} as a new sort and that
$\phi[s,t]$ is short for $\Holds(\phi,s,t)$, where \Holds\ is appropriately
axiomatized.

To see how actions are forced to happen as soon as possible,
let $\ltp(\phi,s,t)$ be an abbreviation for the formula 


$\phi[s,t]\land
t\ge\start(s) \land \forall t'. \start(s)\le t'\ < t\limp
\lnot\phi[s,t']$, 


\noindent that is, $t$ in $\ltp(\phi,s,t)$ is the least time point
after the start of $s$ where $\phi$ becomes true.\footnote{This is not unlike
Reiter's definition of a least natural time point in the context of natural
actions~\cite{Rei96Nat}.  Similar ideas occur in the context of delaying
processes in real-time programming languages like Ada \cite{Bur91Rea}.}

Then we require that a \waitFor-action is possible iff
the condition has a least time point:

\begin{itemize}
\item[] $\Poss(\waitFor(\phi),s) \equiv \exists t.\ltp(\phi,s,t).$
\end{itemize}


It is not hard to show that, if $\exists t.\ltp(\phi,s,t)$ is satisfied, then
$t$ is unique. Finally, we need to characterize how the fluent \start\ changes
its value when an action occurs. The following successor state axiom for
\start\ captures the intuituion that the starting time of a situation changes
only as a result of a $\waitFor(\phi)$, in which case it advances to the
earliest time in the future when $\phi$ holds.

\begin{quote}
  \begin{tabbing}
    \quad \= \quad \= \kill
    $Poss(a,s) \limp [\start(do(a,s)) = t \equiv$\\
    \>$\exists \phi.a = \waitFor(\phi) \land \ltp(\phi,s,t)\lor$\\
    \>$[\forall \phi. a\ne \waitFor(\phi) \land t = \start(s)]]$.
\end{tabbing}
\end{quote}

Let \AX\ be the set of foundational axioms of the previous section together with the
domain closure and unique names axioms for \tfunction{s}, 
the axioms required for \tform's, the precondition
axiom for \waitFor, and the successor state axiom for \start.  Then the
following formulas are logical consequences of $AX$.

\begin{proposition}
\begin{enumerate}
\item The starting time of legal action sequences is monotonically nondecreasing:\\
$\forall s,s'.s \prec s' \limp \start(s) \le \start(s')$.
\item Actions happen as soon as possible:\\
$[\forall a,s.\start(\dofn(a,s)) = \start(s)] \lor [\exists\phi. a = \waitFor(\phi)
\land \ltp(\phi, s, \start(\dofn(a,s)))]$
\end{enumerate}
\end{proposition}

To illustrate the approach, let us go back to the robot example.
First, we can formulate a successor state axiom for \robotLoc:

  \begin{tabbing}
    \quad \= \quad \= \kill
    $Poss(a,s) \limp [\robotLoc(do(a,s)) = y \equiv$\\
    \> $\exists t_0,\vel,x. x = \val(\robotLoc(s),t_0) \land t_0= \start(s) \land$\\
    \> $[a=\startGo(\vel) \land 
        y = \linear(x, \vel, t_0)$\\
    \> \> $\lor a=\stopGo \land y = \constant(x)\lor y = \robotLoc(s) \land$\\
    \> \> $\neg \exists \vel. (a=\startGo(\vel) \lor a=\stopGo)]]$
\end{tabbing}

In other words, when an action is performed \robotLoc\ is assigned either the function
$\linear(x,\vel,t_0)$, if the robot starts moving with velocity \vel\ and
$x$ is the location of the robot at situation $s$, or it is assigned
$\constant(x)$ if the robot stops, or it remains the same as 
in $s$.\footnote{Note that $\val(\robotLoc(s),t_0)$ is well-defined since
every \tfunction\ has a name ($\constant$ or $\linear$) with
corresponding axioms for \val\ as given above.}

Let $\Sigma$ be \AX\ together with the axioms for \val, the
successor state axiom for \robotLoc, precondition axioms
stating that \startGo\ and \stopGo\ are always possible, and the fact
$(\robotLoc(\Szero)=\constant(0))$,
that is, the robot initially rests at position $0$. Let us assume the
robot starts moving at speed 50 (cm/s) and then waits until it reaches
location 1000 (cm), at which point it stops. The resulting situation is
$s_1 = \dofn(\stopGo,\dofn(\waitFor(\robotLoc=1000),\dofn(startGo(50),S_0)))$.
Then
\begin{enumerate}
\item[] $\Sigma\models\start(s_1)=20\land\robotLoc(s_1)=\constant(1000)$.
\end{enumerate}
In other words, the robot moves for 20 seconds and stops at location 1000,
as one would expect.

In summary, to model continuous change and time in the situation calculus,
we have added four new sorts: real, time, \tfunction\ (functions of time), and 
\tform\ (temporal formulas). In addition, we introduced
a special function \val\ to evaluate \tfunction{s}, 
a new kind of primitive action \waitFor\ together
with a domain-independent precondition axiom, and a new fluent \start\ 
(the starting time of a situation) together with a successor state axiom.

\section{\congolog}

\congolog\ \cite{Gia99Fou}, an extension of \golog\ \cite{Lev97GOL}, is
a formalism for specifying complex actions and how these are mapped to
sequences of atomic actions assuming a description of the initial
state of the world, action precondition axioms and successor state
axioms for each fluent.  Complex actions are defined using control
structures familiar from conventional programming language such as
sequence, while-loops and recursive procedures. In addition, parallel
actions are introduced with a conventional interleaving semantics.
Here we confine ourselves to the deterministic fragment of 
\congolog. (While nondeterministic actions raise interesting issues,
we ignore them for reasons of space. Also note that
nondeterminism plays little if any role in languages like \rpl.)

\begin{center}
\begin{tabular}{lr}
$\alpha$ & primitive action\\
$\phi?$ & test action\footnotemark\\
$seq(\sigma_1,\sigma_2)$ & sequence\\
$if(\phi,\sigma_1,\sigma_2)$ & conditional\\
$while(\phi,\sigma)$ & loop\\
$par(\sigma_1, \sigma_2)$ & concurrent execution\\
$prio(\sigma_1, \sigma_2)$ & prioritized execution\\
proc $\beta(x) \sigma$ & procedure definition
\end{tabular}
\end{center}
\footnotetext{Here, $\phi$ stands for a situation calculus formula with all
  situation arguments suppressed, for example $hasMail(gerhard)$.  $\phi[s]$
  will denote the formula obtained by restoring situation variable $s$ to all
  fluents appearing in $\phi$.}


The semantics of \congolog\ is defined using the so-called transition
semantics, which defines single steps of computation. There is a relation,
denoted by the predicate $Trans(\sigma,s,\delta,s')$, that associates with a
given program $\sigma$ and situation $s$ a new situation $s'$ that results
from executing a primitive action in $s$, and a new program $\delta$ that
represents what remains of $\sigma$ after having performed that action.
Furthermore, we need to define which configurations $(\sigma,s)$ are final,
meaning that the computation can be considered completed when a final
configuration is reached.  This is denoted by the predicate
$Final(\sigma,s)$.\footnote{Again, we gloss over the issue of reifying
  formulas and programs in the logical language and refer to~\cite{Gia99Fou}
  for details.}.

We do not consider the axiomatization concerning the \emph{proc} instruction,
which is more subtle to handle.\footnote{Indeed, it necessitates a second
  order axiomatization of $Trans$; see \cite{Gia99Fou} for details.} Note that
the semantics is defined for the {\em non-temporal} situation calculus.
Adapting the semantics to the temporal situation calculus of the previous
section will be the subject of the next section.

\begin{quote}
  \begin{tabbing}
    \quad \= \quad \= \kill
  $Final(\alpha,s) \equiv \false$ , where $\alpha$ is a primitive action\\[1ex]
  $Final(nil,s) \equiv \true$, where $nil$ is the empty program\\[1ex]
  $Final(\phi?,s) \equiv \false$\\[1ex]
  $Final(seq(\sigma_1, \sigma_2),s) \equiv Final(\sigma_1,s) \wedge Final(\sigma_2,s)$\\[1ex]
  $Final(if(\phi,\sigma_1, \sigma_2),s) \equiv$\\
  \> $\phi[s]\land
  Final(\sigma_1,s) \lor \neg \phi[s] \land Final(\sigma_2,s)$\\[1ex]
  $Final(while(\phi,\sigma),s) \equiv \neg \phi[s] \lor Final(\sigma,s)$\\[1ex]
  $Final(par(\sigma_1, \sigma_2),s) \equiv Final(\sigma_1,s) \wedge
  Final(\sigma_2,s)$\\[1ex]
  $Final(prio(\sigma_1, \sigma_2),s) \equiv Final(\sigma_1,s) \wedge
  Final(\sigma_2,s)$
\end{tabbing}
\begin{tabbing}
  \quad \= \quad \= \kill
                                
  $Trans(\alpha,s,\delta,s') \equiv$\\
  \> $\Poss(\alpha,s)\land\delta=nil \land s'=
  \dofn(\alpha,s)$\\[1ex]
  $Trans(nil,s,\delta,s') \equiv \false$\\[1ex]
  $Trans(\phi?,s,\delta,s') \equiv \phi[s] \land \delta = nil \land s' = s$\\[1ex]
  $Trans(seq(\sigma_1,\sigma_2),s,\delta,s') \equiv$\\
  \> $Final(\sigma_1,s) \land Trans(\sigma_2,s,\delta,s') \lor$\\
  \> $\exists \delta' . Trans(\sigma_1,s,\delta',s') \land \delta = seq(\delta',\sigma_2)$\\[1ex]
  $Trans(if(\phi,\sigma_1,\sigma_2),s,\delta,s') \equiv$\\
  \> $\phi[s] \land Trans(\sigma_1,s,\delta,s') \lor$\\
  \> $\neg \phi[s] \land Trans(\sigma_2,s,\delta,s')$\\[1ex]

  $Trans(while(\phi,\sigma),s,\delta,s') \equiv$\\
  \> $\exists \gamma . \delta = seq(\gamma, while(\phi,\sigma)) \land
  \phi[s] \land$\\
  \> \> $ Trans(\sigma,s,\gamma,s')$\\[1ex]

  $Trans(par(\sigma_1,\sigma_2),s,\delta,s') \equiv$\\
  \> $\exists\gamma.\delta=par(\gamma,\sigma_2) \land Trans(\sigma_1,s,\gamma,s')
  \lor$\\
  \> $\exists\gamma.\delta=par(\sigma_1,\gamma) \land Trans(\sigma_2,s,\gamma,s')$\\[1ex]

  $Trans(prio(\sigma_1,\sigma_2),s,\delta,s') \equiv$\\
  \> $\exists\gamma.\delta=prio(\gamma,\sigma_2) \land Trans(\sigma_1,s,\gamma,s')
  \lor$\\
  \> $\exists\gamma.\delta=prio(\sigma_1,\gamma) \land
  Trans(\sigma_2,s,\gamma,s') \land$\\
  \> \> $\not \exists \gamma',s'' . Trans(\sigma_1,s,\gamma',s'')$
  \end{tabbing}
\end{quote}

Intuitively, a program cannot be in its final state if there is still a
primitive action to be done. 
Similarly, a con\-cur\-rent execution of two programs
is in its final state if both are. 
As for $Trans$, let us just look at $par$:
a transition of two programs working in parallel means that one
action of one of the programs is performed.

A final situation $s'$ reachable after a finite number of transitions from
a starting situation is identified with the situation resulting from a
possible execution trace of program $\sigma$, starting in situation $s$; this
is captured by the predicate $Do(\sigma,s,s')$, which is defined in terms of
$Trans^*$, the transitive closure of $Trans$:

\begin{quote}
$Do(\delta,s,s') \equiv \exists \delta'. Trans^*(\delta,s,\delta',s') \land
Final(\delta',s')$

$Trans^*(\delta,s,\delta',s') \equiv \forall T [... \limp
T(\delta,s,\delta',s')]$

\end{quote}


where the ellipsis stands for the conjuction of the 
following formulas:

\begin{quote}
$T(\delta,s,\delta,s)$\\
$Trans(\delta,s,\delta'',s'') \land T(\delta'',s'',\delta',s') \limp
T(\delta,s,\delta',s')$
\end{quote}

Given a program $\delta$, proving that $\delta$ is executable in the initial
situation then amounts to proving $\Sigma\models\exists
s\Do(\delta,S_0,s)$, where $\Sigma$ consists of the above axioms for
\congolog\ together with a basic action theory in the situation calculus.
 

\section{\ccgolog: a Continuous, Concurrent Golog}

Let us now turn to \ccgolog, which is a variant of deterministic \congolog\ 
and which is founded on our new extension of the situation calculus.

First, for reasons discussed below we slightly change the language by replacing
the instructions $par$ and $prio$ by the constructs $tryAll$ and $withPol$,
respectively.  Intuitively, $tryAll(\sigma_1, \sigma_2)$ starts executing both
$\sigma_1$ and $\sigma_2$; but unlike $par$, which requires both
$\sigma_1$ and $\sigma_2$ to reach a final state,
the parallel execution of $tryAll$ stops as soon
as {\em one} of them reaches a final state.
As for $withPol(\sigma_1, \sigma_2)$, the idea is that a low priority plan
$\sigma_2$ is executed, which is interrupted whenever the program $\sigma_1$,
which is called a \emph{policy}, is able to execute. The execution of the whole
$withPol$ construct ends as soon as $\sigma_2$ ends.  (Note that $prio$ is just
like $withPol$ except that for $prio$ to end both $\sigma_1$ and $\sigma_2$ need
to have ended.)

$tryAll$ and $withPol$ are inspired by
similar instructions in \rpl\ where they have been found very useful in specifying
complex concurrent behavior. In particular, $withPol$ is useful
to specify the execution of a plan while guarding certain constraints. 
As we will see later, it is quite straightforward to define $par$ and $prio$
using the new instructions. On the other hand, defining $tryAll$ and $withPol$ in terms
of $par$ and $prio$ appears to be more complicated. Hence we decided to
trade $par$ and $prio$ for their siblings.


Let us now turn to the semantics of \ccgolog, which means finding appropriate
definitions for $Final$ and $Trans$. To start with, the semantics remains
exactly the same for all those constructs inherited from deterministic
\congolog. Note that this is also true for the new $\waitFor(\phi)$, which is
treated like any other primitive action.\footnote{The reader familiar with
  \congolog\ may wonder whether a test action $\phi?$ is the same as
  $\waitFor(\phi)$. This is not so. Roughly, the main difference is that tests
  have no effect on the world while \waitFor\ advances the time.}  Hence we
are left to deal with $tryAll$ and $withPol$.

It is straightforward to give $Final$ its intended meaning, that is,
$tryAll$ ends if one of the two programs ends and $withPol$ ends
if the second program ends:\\[-9pt]

%

\indent  $Final(tryAll(\sigma_1, \sigma_2,s)) \equiv  Final(\sigma_1,s) \lor Final(\sigma_2,s)$\\
\indent  $Final(withPol(\sigma_1, \sigma_2),s) \equiv Final(\sigma_2,s)$

\vspace*{1pt}
When considering the transition of concurrent programs, care must be taken in
order to avoid conflicts with the assumption that actions should happen as soon
as possible, which underlies our new version of the situation calculus. To see
why let us consider the following example, where we want to instruct our robot
to run a backup at time 8 or 20, whichever comes first. Let us assume we have a
continuous fluent $\clock$ representing the time\footnote{This can be
modeled using a simple linear function, but we ignore the details here.} and let
$\runBackup$ be always possible. Given our intuitive reading of $tryAll$, we may
want to use the following program:

\begin{tabbing}
  \indent  $seq($\=$tryAll(\waitFor(\clock = 8),\waitFor(\clock = 20)),$\\
  \> $\runBackup)$
\end{tabbing}

\noindent If we start the program at time 0 we would expect to see

\begin{itemize}
\item[] [\waitFor(\clock=8),\runBackup]
\end{itemize}

\noindent as the only execution trace, since time 8 is reached first. (Recall that
$tryAll$ finishes as soon as one of its arguments finishes.)
However, this is not necessarily guaranteed. In fact,
the obvious adaptation of \congolog's $Trans$-definition of $par$ to the
case of $tryAll$\footnote{Roughly, replace $par$ by $tryAll$ and
add $\neg Final(\sigma_1,s)$ and $\neg Final(\sigma_2,s)$ as additional
conjuncts on the definition's R.H.S.} also yields the trace

\vspace*{-3pt}
\begin{itemize}
\item[] [\waitFor(\clock=20),\runBackup].
\end{itemize}
\vspace*{-3pt}

\noindent This is because there simply is no preference enforced between the
two \waitFor-actions. As the following definition shows, it is
not hard to require that actions which can be executed earlier
are always preferred, restoring the original idea that actions should
happen as early as possible.

  \begin{tabbing}
    \quad \= \quad \= \kill
    $Trans(tryAll(\sigma_1,\sigma_2),s,\delta,s') \equiv$\\

    \> $\neg Final(\sigma_1,s) \land \neg Final(\sigma_2,s) \land$\\
    \> \> $\exists \delta_1 . Trans(\sigma_1,s,\delta_1,s') \land
    \delta=tryAll(\delta_1,\sigma_2) \land$\\
    \> \>$\forall \delta_2,s_2.Trans(\sigma_2,s,\delta_2,s_2) \limp
start(s')
    \leq start(s_2)]$\\
    \> $\lor$ \>$\exists \delta_2 . Trans(\sigma_2,s,\delta_2,s') \land
    \delta=tryAll(\sigma_1,\delta_2) \land$\\
    \> \>$\forall \delta_1,s_1.Trans(\sigma_1,s,\delta_1,s_1) \limp
start(s')
    \leq start(s_1)]$
  \end{tabbing}

We are left with defining $Trans$ for $withPol$. To see what is involved,
let us consider the following example
\begin{quote}
  $withPol(watchB,deliverMail)$, where
\end{quote}
$watchB = seq(\waitFor(battLevel\le 46),chargeBatt)$.\\ The idea is to deliver
mail and, with higher priority, watch for a low battery level, at which point
the batteries are charged.  In the discussion of a similar scenario written in
\rpl, we already pointed out that the \waitFor-action should not block the mail
delivery even though it belongs to the high priority policy. On the other hand,
once the routine for charging the batteries starts, it should not be interrupted,
that is, it should run in blocking mode, which should also hold for
possible \waitFor-actions it may contain such as waiting for
arrival at the docking station. It turns out that it suffices to arrange in the
semantics of $Trans$ that occurrences of \waitFor\ within a policy are
considered non-blocking. As we will see below, the effect of a policy
running in blocking mode is definable by other means.

Interestingly, the resulting axiom is almost identical to that of
$tryAll$: the main difference is that $\leq$ is replaced by $<$ in the last line.
This ensures that $\sigma_1$ takes precedence if both $\sigma_i$ are about to
execute an action at the same time.

  \begin{tabbing}
    \quad \= \quad \= \kill
    $Trans(withPol(\sigma_1,\sigma_2),s,\delta,s') \equiv \neg Final(\sigma_2,s) \land$\\
    \> \> $\exists \delta_1 . Trans(\sigma_1,s,\delta_1,s') \land
    \delta=withPol(\delta_1,\sigma_2) \land$\\
    \> \>$\forall \delta_2,s_2.Trans(\sigma_2,s,\delta_2,s_2) \limp
start(s')
    \leq start(s_2)]$\\
    \> $\lor$ \>$\exists \delta_2 . Trans(\sigma_2,s,\delta_2,s') \land
    \delta=withPol(\sigma_1,\delta_2) \land$\\
    \> \>$\forall \delta_1,s_1.Trans(\sigma_1,s,\delta_1,s_1) \limp
start(s')
    < start(s_1)]$

  \end{tabbing}

This then ends the discussion of the semantics of $Trans$ in \ccgolog.
$Trans^*$ and $Do(\delta,s,s')$ are defined the same way as in \congolog.\\[-2ex]

One issue left open is to show how a policy can run in blocking mode. This can
be arranged using the macro $withCtrl(\phi,\sigma)$, which stands for $\sigma$
with every primitive action or test $\alpha$ replaced by
$if(\phi,\alpha,\false?)$.\footnote{We remark that $if(\phi,\alpha,\false?)$
  can only lead to a transition if $\phi$ is true in the current situation at
  which point $\alpha$ is executed immediately.  This is essentially due to
  the fact that $\false?$ is neither $Final$ nor can it ever lead to a
  transition (see~\cite{Gia99Fou}).}

Intuitively $withCtrl(\phi,\sigma)$ executes $\sigma$ as long as $\phi$ is
true, but gets blocked otherwise. As the following example shows, the effect
of a policy in blocking mode is obtained by having the truth value of $\phi$
be controlled by the policy and using the $withCtrl(\phi,\sigma)$-construct in
the low priority program.

This leads us, finally, to the specification of our initial example in
\ccgolog. In the following we assume a fluent $wheels$, which is initially
\true, set \false\ by $grabWhls$, and reset by the action $releaseWhls$.  We
also use $whenever(\phi,\sigma)$ as shorthand for
$while(true,seq(waitFor(\phi),\sigma))$.

\begin{figure}[htbp]
\shadowbox{
  \begin{minipage}{5cm}

\begin{tabbing}
  \quad \= \quad \= \quad \= \kill
  $withPol(whenever(battLevel \leq 46,$\\
  \> \> $seq(grabWhls,\chargeBatteries,releaseWhls)$\footnotemark$),$\\
  \> $withPol(whenever(nearDoor\mbox{A-118},$\\
  \> \> \> $seq(\say(hello),\waitFor(\neg nearDoor\mbox{A-118})))$\\
  \> \> $withCtrl(wheels,deliverMail)))$
\end{tabbing}
  \end{minipage}
  } 
  \caption{The introductory example as \ccgolog\ plan.}
  \label{fig:introex}
\end{figure}

\footnotetext{ $seq(\sigma_1,\sigma_2,\sigma_3)$ is a shorthand for
  $seq(\sigma_1,seq(\sigma_2,\sigma_3))$. We will also use a similar shorthand
  for $tryAll$.}

In this program, the outermost policy is waiting until the battery level drops
to $46$. At this point, a $grabWheels$ is immediately executed, which blocks
the execution of the program $deliverMail$. It is only after the complete
execution of $\chargeBatteries$ that $wheels$ gets released so that
$deliverMail$ may resume execution (if, while driving to the batterie docking
station, the robot passes by $RmA-118$, it would still say ``hello'').

Note that the \ccgolog-program is in a form very close to the original
\rpl-program we started out with.  Hence we feel that \ccgolog\ is a step in the
right direction towards modeling more realistic domains which so far could only
be dealt with in non-logic-based approaches. Moreover, with their rigorous
logical foundation, it is now possible to make provable predictions about how
the world evolves when executing \ccgolog-programs. (See also the next section
on experimental results.)

Finally, let us briefly consider how $par$ and $prio$ which we dropped in favor
of $tryAll$ and $withPol$ are definable within \ccgolog.  Let us assume fluents
$flg_i$ which are initially \false\ and set \true\ by $setFlg_i$. Then we can
achieve what amounts to $par(\sigma_1,\sigma_2)$ by
$tryAll(seq(\sigma_1,setFlg1,flg2?),seq(\sigma_2,setFlg2,flg1?))$.  Note that
the testing of the flags at the end of each program forces that both $\sigma_i$
need to finish.  Similarly, $prio$ can be defined as
$withPol(seq(\sigma_1,setFlg),seq(\sigma_2,flg?))$.

We end this section with some remarks on Reiter's proposal for a temporal
version of \golog~\cite{Rei98Seq},\footnote{While the paper is about
  sequential \golog, the extension to \congolog\ is straightforward.}  which
makes use of a different temporal extension of the situation
calculus~\cite{Pin97Int,Rei96Nat}. Roughly, the idea is that every primitive
action has as an extra argument its execution time. E.g., we would write
$\stopGo(20)$ to indicate that \stopGo\ is executed at time $20$. It turns out
that this explicit mention of time is highly problematic when it comes to
formulating programs such as the above.  Consider the part about saying
``hello'' whenever the robot is near Room A-118. In Reiter's approach, the
programmer would have to supply a temporal expression as an argument of the
\say-action. However, it is far from obvious what this expression would look
like since it involves analyzing the mail delivery subprogram as well as
considering the odd chance of a battery recharge. In a nutshell, while
Reiter's approach forces the user to figure out when to act, we let \ccgolog\ 
do the work.  --- As a final aside, we remark that \waitFor-actions allow us
to easily emulate Reiter's approach within our framework.

\section{Experimental Results}
Although the definition of \ccgolog\ requires second-order logic, it is easy
to implement a PROLOG interpreter for \ccgolog, just as in the case of the
original \congolog.\footnote{The subtle differences between the second order
  axiomatization of \congolog\ and a PROLOG implementation are discussed in
  \cite{Gia99Fou}.}  In order to deal with the constraints implied by the
$waitFor$ instruction, we have made use of the ECRC Common Logic Programming
System Eclipse 4.2 and its built-in constraint solver library \texttt{clpr} to
implement a \ccgolog\ interpreter (similar to Reiter \cite{Rei98Seq}).

\begin{figure}

  \vspace*{2ex}
    \begin{center}
      \setlength{\epsfxsize}{.75\columnwidth}
      \epsffile{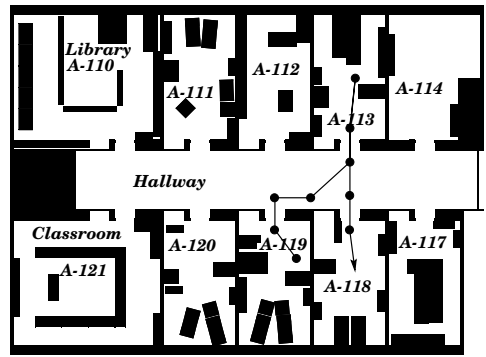}
    \end{center}

  \caption{The robot environment}
  \label{fig:env}

\end{figure}

We used three example plans to compare the performance of our \ccgolog\ 
interpreter with earlier work on the projection of continuous change
\cite{Bee98Cau} within the \xfrm\ \cite{McD92Rob,McD94Alg} framework. As an
example environment, we use the one of \cite{Bee98Cau}, depicted in
Figure~\ref{fig:env}. We approximated the robot's trajectory by polylines,
consisting of the starting location, the goal location and a point in front of
and behind every passed doorway (similar to \cite{Bee98Cau}):

\begin{itemize}
\item The introductory example (Fig.~\ref{fig:introex}). We assumed that there
  are two letters to be delivered and that the delivery is once interrupted by
  low battery voltage.
\item The (slightly modified) example of \cite{Bee98Cau}. Again, the robot is
  to deliver letters. But at the same time, it has to monitor the state of the
  environment, that is it has to check wether doors are open. As soon as it
  realizes that the door to A-113 is open, it has to interrupt its actual
  delivery in order to deliver an urgent letter to A-113. This is specified as
  a policy that leads the robot inside A-113 as soon as the opportunity is
  recognizes. Note that in the implementation, we used PROLOG lists
  (\texttt{[a1,a2,...]}) instead of $seq(a_1,a_2,...)$.

  \begin{small} \begin{verbatim}
withPol(whenever(inHallway,
  [say(enterHW),
    tryAll([whenever(nearDoor(a-110),
              [checkDoor(a-110+121),false?])],
           [whenever(nearDoor(a-111),
              [checkDoor(a-111+120),false?])],
           [whenever(nearDoor(a-112),
              [checkDoor(a-112+119),false?])],
           [whenever(nearDoor(a-113),
              [checkDoor(a-113+118),false?])],
           [whenever(nearDoor(a-114),
              [checkDoor(a-114+117),false?])],
           [waitFor(leftHallway),
              say(leftHW)])]),
   withPol([useOpp?,
            gotoRoom(a-113),deliverUrgentMail],
      [gotoRoom(a-118),giveMail(gerhard)]))).
  \end{verbatim}
  \end{small}

  
  The outer policy is activated whenever the robot enters the hallway. It
  concurrently monitors whether the robot reaches a location near a door,
  whereat is checks whether the door is open or not. If A-113 is detected to
  be open, the fluent \texttt{useOpp} is set true (by procudure
  \texttt{checkDoor}). The policy is deactivated when the robot leaves the
  hallway.
  
  The inner policy is activated as soon as \texttt{useOpp} gets true. It's
  purpose is to use the opportunity to enter A-113 as soon as possible.
  Figure~\ref{fig:env} illustrates the projected trajectory, assuming that the
  door to A-113 is open.

\item A longer trajectory through all rooms.

  \begin{small}

\begin{verbatim}
withPol(whenever(inHallway,
  [say(enterHW),
  [say(enterHW),
    tryAll([whenever(nearDoor(a-110),
              [checkDoor(a-110+121),false?])],
           ...]),
      [gotoRoom(a-114),say(a-114),
       gotoRoom(a-113),say(a-113),
       gotoRoom(a-112),say(a-112),
       gotoRoom(a-111),say(a-111),
       gotoRoom(a-110),say(a-110),
       gotoRoom(a-117),say(a-117),
       gotoRoom(a-118),say(a-118),
       gotoRoom(a-119),say(a-119),
       gotoRoom(a-120),say(a-120),
       gotoRoom(a-121),say(a-121)
      ])))))
  \end{verbatim}
  \end{small}

\end{itemize}

Figure~\ref{fig:runtimes} shows the time it took to generate a projection of
the example plans using \ccgolog\ resp. \xfrm, as well as the number of
actions resp.  events occuring in the projection. Both \ccgolog\ and \xfrm\ 
run on the same machine (a Linux Pentium III Workstation), under Allegro
Common Lisp Trial Edition 5.0 resp.  Eclipse 4.2.  As it turns out, \ccgolog\ 
appears to be faster by an order of magnitude than \xfrm.  We believe that
\ccgolog\ owes this somewhat surprising advantage to the fact that it lends
itself to a simple implementation with little overhead, while \xfrm\ relies on
the rather complex \rpl-interpreter involving many thousand lines of Lisp
code. 

\begin{figure}[htbp]
  \begin{center}
    \begin{tabular}{|l|l|l|l|l|l|}
      \hline
      Problem:    &  \ccgolog  & \xfrm\\
      \hline
      \hline
      {\sc Intro. Ex.} & 0.4 s / 115 acts & - \\
      \hline
      {\sc AIPS-98. Ex.} & 0.5 s / 73 acts & 3.6 s / 106 evs\\
      \hline
      {\sc Long Traj.} & 3 s / 355 evs & 22.7 / 486 evs\\
      \hline
    \end{tabular}
  \end{center}
  \caption{Runtime in seconds}
  \label{fig:runtimes}
\end{figure}

It also seems noteworthy that \ccgolog\ contents itself with significantly
less predicted relevant events (i.e. atomic actions) than \xfrm. This results
from the fact that \ccgolog\ only predicts an action if an action is actually
executed (unlike \xfrm, which also projects events whose only purpose is to
test if actual changes occur, like the ``reschedule'' events; see
\cite{Bee98Cau}). Finally, and maybe most importantly, the \ccgolog\ 
implementation is firmly based on a logical specification, while \xfrm\ relies
on the procedural semantics of the \rpl\ interpreter.

\section{Conclusions}

In this paper we proposed an extension of the situation calculus which
includes a model of continuous change due to Pinto and a novel approach to
modeling the passage of time using a special \waitFor-action.  We then
considered \ccgolog, a deterministic variant of \congolog\ which is based on
the extended situation calculus.  A key feature of the new language is the
ability to have part of a program wait for an event like the battery voltage
dropping dangerously low while other parts of the program run in parallel.
Such mechanisms allow very natural formulations of robot controllers, in
particular, because there is no need to state explicitly in the program when
actions should occur. In addition to the sound theoretical foundations on
which \ccgolog\ is built, experimental results have shown a superior
performance in computing projections when compared to the projection mechanism
of the plan language \rpl, whose expressive power has largely motivated the
development of \ccgolog.

However, much remains to be done. For one, sensing actions need to be properly
integrated into this approach. Here we hope to benefit from existing approaches
in \golog\ and \congolog~\cite{Lak99CAT,Gia98Inc}. Also, uncertainty plays
a central role in the robotics domain which should be reflected in
a plan language as well. Based on foundational work within the
situation calculus~\cite{Bac95Noi} first preliminary results have been obtained
regarding an integration into \congolog~\cite{Gro99Pro,Gro00Tur}.

Finally, a few words are in order regarding the use of projections in
\ccgolog. They should be understood as a way of assessing whether a program is
executable {\em in principle}. The resulting execution trace of a projection is
not intended as input to the execution mechanism of the robot.  This is because
the time point of a \waitFor-condition like a low battery level is computed
based on a {\em model} of the world which includes a model of the robot's energy
consumption. In reality, of course, the robot should react to the {\em actual}
battery level by periodically reading its voltage meter. In the runtime system
of \rpl\ for an actual robot \cite{Thr99Min} this link between \waitFor-actions
and basic sensors which are immediately accessible to the robot has been
realized. One possibility to actually execute \ccgolog-programs on a robot would
be to combine this idea of executing \waitFor's with an incremental interpreter
along the lines of~\cite{Gia99Inc}. We leave this to future work.
  
\bibliography{publications}

\end{document}